\documentclass{article}
\usepackage{times}
\usepackage{soul}
\usepackage{url}
\usepackage[hidelinks]{hyperref}
\usepackage[utf8]{inputenc}
\usepackage[small]{caption}
\usepackage{graphicx}
\usepackage{amsmath}
\usepackage{booktabs}
\usepackage{helvet}  %Required
\usepackage{courier}  %Required
\usepackage{amsmath,amssymb, amsthm}
\usepackage{latexsym}
\usepackage{color}
\usepackage{tabularx}
\usepackage[english]{babel}
\usepackage{algorithmicx, algpseudocode, algorithm}
\usepackage{mathrsfs}
\usepackage{bm}
\usepackage{tikz}
\usepackage{enumerate}

\newtheorem{Theorem}{Theorem}

\newtheorem{Proposition}[Theorem]{Proposition}

\newtheorem{Assumption}{Assumption}

\newtheorem{Definition}{Definition}
\newenvironment{sproof}{%
  \proof}{\endproof}

\renewcommand{\algorithmicrequire}{\textbf{Input:}}
\renewcommand{\algorithmicensure}{\textbf{Output:}}

\newcommand{\betahat}{{\hat{\beta}}}

\newcommand{\argmin}{\mathop{\rm argmin}\limits}

\newcommand{\Sigmatilde}{{\tilde{\Sigma}}}
\newcommand{\Sigmahat}{\hat{\Sigma}}
\newcommand{\Sigmastar}{\Sigma^*}
\newcommand{\Sigmacheck}{\check{\Sigma}}
\newcommand{\Spair}{S^{\rm pair}}

\newcommand{\Simp}{S^{\rm imp}}

\newcommand{\rhopair}{\rho^{\rm pair}}

\newcommand{\sgn}{{\rm sgn}}
\newcommand{\vecl}{{\rm vec}}
\newcommand{\matl}{{\rm mat}}

\title{HMLasso: Lasso with High Missing Rate}

\author{Masaaki Takada${}^{1}$, Hironori Fujisawa${}^{2}$, Takeichiro Nishikawa${}^{1}$\\
${}^{1}$ Toshiba Corporation\\
${}^{2}$ The Institute of Statistical Mathematics
}
\date{}

\begin{document}

\maketitle

\begin{abstract}

Sparse regression such as the Lasso has achieved great success in handling high-dimensional data. However, one of the biggest practical problems is that high-dimensional data often contain large amounts of missing values. Convex Conditioned Lasso (CoCoLasso) has been proposed for dealing with high-dimensional data with missing values, but it performs poorly when there are many missing values, so that the high missing rate problem has not been resolved. In this paper, we propose a novel Lasso-type regression method for high-dimensional data with high missing rates. We effectively incorporate mean imputed covariance, overcoming its inherent estimation bias. The result is an optimally weighted modification of CoCoLasso according to missing ratios. We theoretically and experimentally show that our proposed method is highly effective even when there are many missing values.

\end{abstract}

\section{Introduction}

High-dimensional data appear in a wide range of fields, including biology, economy, and industry.
Over the past several decades, sparse regression has achieved great success in dealing with high-dimensional data, because it efficiently performs both model estimation and variable selection simultaneously.
Sparse regression methods include the Lasso~\cite{Tibshirani:1996}, Generalized Lasso~\cite{Tibshirani:2011}, Elastic Net~\cite{Zou:2005}, SCAD~\cite{Zou:2001}, and MCP~\cite{Zhang:2010}.

In practice, high-dimensional data often contain large amounts of missing values.
For example, educational and psychological studies commonly have missing data ratios of 15--20\%~\cite{Enders:2003}, while maintenance data for typical industrial processes had over 75\% missing values in over 50\% of variables~\cite{Lakshminarayan:1999}.
Up to 90\% of traffic data can be missing~\cite{Tan:2013}.
There is thus demand for methods that can accommodate high-dimensional data with a high missing rate.

Missing data analysis has a long history.
Listwise deletion (complete case analysis) and pairwise deletion are widely used because of their simplicity.
Various methods have been developed,
including the expectation maximization (EM) algorithm~\cite{Dempster:1977,Schafer:1997}, multiple imputation (MI)~\cite{Little:1987,Little:2002,Schafer:1997,Buuren:2011}, and full information maximum likelihood (FIML)~\cite{Hartley:1971,Enders:2001}.
However, these methods focus on low-dimensional missing data and are intractable for high-dimensional data due to their computational cost.

To deal with high-dimensional missing data, a direct regression method using a pairwise covariance matrix has been proposed~\cite{Loh:2012}.
This method incurs low computational costs, but heavily depends on some critical unknown parameters that must be determined in advance due to its nonconvexity.
Convex conditioned Lasso (CoCoLasso) was proposed to avoid this problem~\cite{Datta:2017}.
Because of its convexity using a positive semidefinite approximation of the pairwise covariance matrix, it does not suffer from local optima or critical parameters.
However, we found that CoCoLasso can deteriorate at high missing rates.
Indeed, estimator accuracy may significantly worsen even when only one variable has a high missing rate, so a high missing rate remains problematic.

In this paper, we propose a novel regression method that overcomes problems related to both high-dimensionality and high missing rates.
We use the mean imputed covariance matrix, effectively incorporating it into Lasso despite its noted tendency toward estimation bias for missing data.
The resulting optimization problem can be seen as a weighted modification of CoCoLasso using the missing ratio, and it is quite effective for high-dimensional data with a high missing rate.
Our proposed method is free from local optima and critical parameters due to convexity, and it is theoretically and experimentally superior to CoCoLasso regarding estimation error.
Contributions of this study are as follows:
\begin{itemize}
  \item We propose a novel regression method for handling high-dimensional data with high missing rates.
  \item We analyze theoretical properties of our method, showing that our formulation
  is superior to all other weighted formulations with regards to estimation error.
  \item We demonstrate the effectiveness of our method through both numerical simulations and real-world data experiments.
  Our method outperforms other methods in almost all cases, and particularly shows significant improvement for high-dimensional data with high missing rates.
\end{itemize}

The remainder of this paper is organized as follows:
We first review existing methods and describe our proposed method.
We then show the advantages of our method through theoretical analyses, numerical simulations, and real-world data experiments.

\subsection*{Notations}
Let $v \in \mathbb{R}^{p}$.

$\|v\|_q~(q>0)$ is the $\ell_q$ norm, that is, $\|v\|_q = \left(|v_1|^q + \dots + |v_p|^q \right)^{1/q}$.
Let $M \in \mathbb{R}^{n \times p}$.
$\| M \|_{\rm F}$ is the Frobenius norm, that is, $\| M \|_{\rm F} = (\sum_{j,k} M_{jk}^2 )^{1/2}$.
$ \| M \|_{\max}$ is the max norm, that is, $\| M \|_{\max} = \max_{j, k} |M_{jk}|$.
Let $M_1, M_2 \in \mathbb{R}^{n\times p}$.
$M_1 \odot M_2$ is the element-wise product (Hadamard product) of $M_1$ and $M_2$.
$M_1 \oslash M_2$ is the element-wise division of $M_1$ and $M_2$.
Let $M \in \mathbb{R}^{p \times p}$ be a symmetric matrix.
$M \succeq 0$ denotes that $M$ is positive semidefinite, that is, $v^\top M v \geq 0$ for any $v \in \mathbb{R}^p$.

\section{Methods}

\subsection{Problem Formulation}
Consider a linear regression model
$
y = X \beta + \varepsilon,
$
where $X \in \mathbb{R}^{n\times p}$ is a complete design matrix,
$y \in \mathbb{R}^{n}$ is a response,
$\beta \in \mathbb{R}^{p}$ is a regression coefficient,
and $\varepsilon \in \mathbb{R}^{n}$ is a noise.
Suppose that $y$ and each column of $X$ are centered without loss of generality.
The ordinary problem is to estimate the regression coefficient $\beta$ given complete data $X$ and $y$.
In contrast, we consider the situation where some elements of $X$ are missing in this paper.

If $X$ does not contain missing values, Lasso is one of the most promising methods for handling high-dimensional data.
It solves the problem~\cite{Tibshirani:1996}
\begin{align}
  \betahat = \argmin_\beta \frac{1}{2n} \| y - X \beta \|_2^2 + \lambda \| \beta \|_1, \label{lasso}
\end{align}
where $\lambda > 0$ is a regularization parameter.
Since the objective function~\eqref{lasso} is regularized by the $\ell_1$ norm of $\beta$, the solution is sparse and often has a small generalization error.
In the presence of missing values, however, it is impossible to directly apply the Lasso.

\subsection{Review of Existing Methods}

Interestingly, to estimate the parameter $\beta$ in the presence of missing values does {\it not} require imputation if $X$, which is computationally expensive for high-dimensional data.
We can directly estimate the parameter without imputation.
The Lasso objective function~\eqref{lasso} can be reformulated as
\begin{align}
  \betahat = \argmin_\beta \frac{1}{2} \beta^\top S \beta - \rho^\top \beta + \lambda \| \beta \|_1, \label{reformulation}
\end{align}
where $S=\frac{1}{n} X^\top X$ (the sample covariance matrix of $X$) and $\rho=\frac{1}{n}X^\top y$ (the sample covariance vector of $X$ and $y$).
Using~\eqref{reformulation}, we can estimate $\beta$ via $S$ and $\rho$ instead of $X$ and $y$.
If data are missing completely at random, we can easily construct unbiased estimators of the covariance matrix and vector using the pairwise covariance, that is, $\Spair=(S_{jk}^{\rm pair})$ and $\rho^{\rm pair}=(\rho_{jk}^{\rm pair})$ as
$$
  S_{jk}^{\rm pair} := \frac{1}{n_{jk}} \sum_{i \in I_{jk}} X_{ij}X_{ik}, \ {\rm and} \
  \rho_j^{\rm pair} := \frac{1}{n_{jj}} \sum_{i \in I_{jj}} X_{ij}y_i,
$$
where $I_{jk} := \{ i : $ $X_{ij}$ and $X_{ik}$ are observed $\}$,
and $n_{jk}$ is the number of elements of $I_{jk}$.
Thus, we can replace $S$ and $\rho$ by $\Spair$ and $\rho^{\rm pair}$ in~\eqref{reformulation}, respectively.

The major problem here is that $\Spair$ may {\it not} be positive semidefinite (PSD). In other words, it may have negative eigenvalues.
This is a critical problem because negative eigenvalues can cause the objective function to diverge to minus infinity, meaning the optimization failed.
A regression method with a nonconvex constrained formulation has been proposed to avoid this problem~\cite{Loh:2012}.
However, this method's nonconvexity causes some difficulties in practice. Namely, different initial values result in multiple global or local minimizers that produce different output solutions, and the solutions depend on unknown parameters that must be determined in advance.
To avoid the above difficulties, a convex optimization problem called CoCoLasso, has been proposed~\cite{Datta:2017}. This problem is formulated as
\begin{align}
  \betahat &= \argmin_\beta \frac{1}{2}\beta^\top\Sigmacheck\beta
	- {\rhopair}^\top\beta + \lambda\|\beta\|_1, \label{cocolasso-betahat}\\
	\Sigmacheck &= \argmin_{\Sigma \succeq 0} \|\Sigma - \Spair\|_{\max}. \label{Sigmacheck}
\end{align}
CoCoLasso obtains the PSD covariance matrix in~\eqref{Sigmacheck} via the alternating direction method of multipliers (ADMM) algorithm, then optimizes the Lasso objective function~\eqref{cocolasso-betahat}.
This formulation overcomes the difficulties seen in~\cite{Loh:2012} because
the objective function~\eqref{cocolasso-betahat} is convex due to the PSD matrix $\Sigmacheck$, meaning it has no local minimizers, and because it uses no unknown parameters that must be determined in advance.
In addition, statistical non-asymptotic properties were also derived.
For these reasons, CoCoLasso is practical and state-of-the art.

However, a high missing rate can deteriorate estimations of the covariance matrix in CoCoLasso.
If some pairwise observation numbers $n_{jk}$ are very small, then the corresponding pairwise covariances $\Spair_{jk}$ are quite unreliable,
possibly becoming very large or small.
Since~\eqref{Sigmacheck} is based on the max norm,
unreliable elements of $\Spair$ will greatly affect the estimator.
As a result, other estimator elements can highly deviate from the corresponding elements in $\Spair$, even if their variables have few missing values.
This indicates that CoCoLasso results can significantly worsen, even if only one variable has a high missing rate.
The problem is that CoCoLasso does not account for the differences in reliability of the pairwise covariance.
The next subsection describes how we overcome this problem.

We mension
other approaches for regression with missing data.
A simple approach is listwise deletion.
This method is very fast but inappropriate when there are few complete samples, as is common with high-dimensional data.
Another typical approach is to impute missing values.
This approach includes various imputation methods, including mean imputation,
the EM algorithm~\cite{Dempster:1977,Schafer:1997}, MI~\cite{Little:1987,Little:2002,Schafer:1997,Buuren:2011}, FIML~\cite{Hartley:1971,Enders:2001}, and other non-parametric methods~\cite{Stekhoven:2012}.
These methods, however, typically incur large computational costs or cause large bias in high-dimensional problems,
and are difficult to conduct theoretical analyses.
Other direct modeling methods use the Dantzig selector instead of the Lasso~\cite{Rosenbaum:2010,Rosenbaum:2013}.
An advantage of the Lasso-type approach is that computation is empirically much faster than with the Dantzig selector~\cite{Efron:2007}.

\subsection{Proposed Method: HMLasso}
The mean imputation method is commonly used in practice.
Let $Z$ be the mean imputed data of $X$.
Because $X$ is centered, $Z_{jk} = X_{jk}$ for observed elements and $Z_{jk} = 0$ for missing elements.
The covariance matrix of the mean imputed data, $S^{\rm imp}=(\Simp_{jk})$, is defined as
\begin{align}
  S_{jk}^{\rm imp} = \frac{1}{n} \sum_{i=1}^n Z_{ij} Z_{ik}
  = \frac{n_{jk}}{n} \frac{1}{n_{jk}} \sum_{i \in I_{jk}} X_{ij} X_{ik}
  = \frac{n_{jk}}{n} S_{jk}^{\rm pair}.
\label{Simppair}
\end{align}
We can equivalently express~\eqref{Simppair} as
\begin{align}
  S^{\rm imp} = R \odot S^{\rm pair} \label{Simppair2}
\end{align}
where $R = (R_{jk})$ with $R_{jk} = n_{jk} / n$.
The mean imputed covariance matrix $\Simp$
is biased but PSD, while the pairwise covariance matrix $\Spair$
is unbiased but not PSD.
To take the best aspects of both, we optimize
\begin{align}
  \Sigmatilde
  &= \argmin_{\Sigma \succeq 0} \| R \odot \Sigma - \Simp \|_{\rm F}^2
  \label{Stilde-Simp}
\end{align}
to obtain a low-biased and PSD matrix.
Direct use of mean imputation for covariance matrix estimation is known to produce estimation bias.
However, we can use the relation~\eqref{Simppair2} between $\Simp$ and $\Spair$ to effectively incorporate them into the optimization problem~\eqref{Stilde-Simp}.

The formulation~\eqref{Stilde-Simp} has a useful property, in that it is equivalent to
\begin{align}
  \Sigmatilde
  &= \argmin_{\Sigma \succeq 0} \| R \odot (\Sigma - \Spair) \|_{\rm F}^2. \label{R-Stilde}
\end{align}
The formulation \eqref{R-Stilde} can be seen as a weighted modification of CoCoLasso~\eqref{Sigmacheck} using the observed ratio matrix $R$, where the max norm is replaced by the Frobenius norm.
This weighting is beneficial under high missing rates.
When there are missing observations, the objective function downweights the corresponding term $\Sigma_{jk}-\Spair_{jk}$ by the observed ratio $R_{jk} = n_{jk} / n$.
In particular, when $n_{jk}$ is small, the downweighting will be reasonable, because the pairwise covariance $\Spair_{jk}$ is unreliable.

From the above, we extend the formulation and propose
a novel optimization problem to estimate the regression model as
\begin{align}
  \betahat &= \argmin_\beta \frac{1}{2}\beta^\top\Sigmatilde\beta
	- {\rhopair}^\top\beta + \lambda\|\beta\|_1, \label{betahat}\\
  \Sigmatilde
  &= \argmin_{\Sigma \succeq 0} \| W \odot (\Sigma - \Spair) \|_{\rm F}^2, \label{Sigmatilde}
\end{align}
where $W$ is a weight matrix whose $(j,k)$-th element is $W_{jk}=R_{jk}^\alpha$ with a constant $\alpha \ge 0$.
We obtains a PSD matrix by minimizing the weighted Frobenius norm in~\eqref{Sigmatilde},
then optimizes the Lasso problem~\eqref{betahat}.
This Lasso-type formulation allows us to efficiently deal with high missing rates, so we call our method ``HMLasso''.

Several $\alpha$ values can be considered.
Setting $\alpha=0$ corresponds to the non-weighted case, which is just a projection of $\Spair$ onto the PSD region.
This is the same as CoCoLasso when the Frobenius norm in~\eqref{Sigmatilde} is replaced by the max norm.
The case where $\alpha=1$ relates to mean imputation, as described above.
As shown below, non-asymptotic analyses support that $\alpha=1$ is reasonable, and numerical experiments show that this setting delivers the best performance.
Therefore, we recommend setting $\alpha=1$ in practice.
The case where $\alpha=1/2$ can be roughly viewed as the maximum likelihood method from an asymptotic perspective.
We discuss the case of setting $\alpha=1/2$ in Appendix \ref{apx:asymptotic}.

Note that in~\eqref{Sigmatilde}, we use the Frobenius norm instead of the max norm,
because the Frobenius norm delivered better performance in numerical experiments.

\subsection{Comparison Using a Simple Example}

\begin{figure}[t]
  % \begin{center}
  \centering
  \begin{tikzpicture}
    % \draw [help lines] (0,0) grid (9,3);
    \node at (3,1.5) {\includegraphics[scale=0.5]{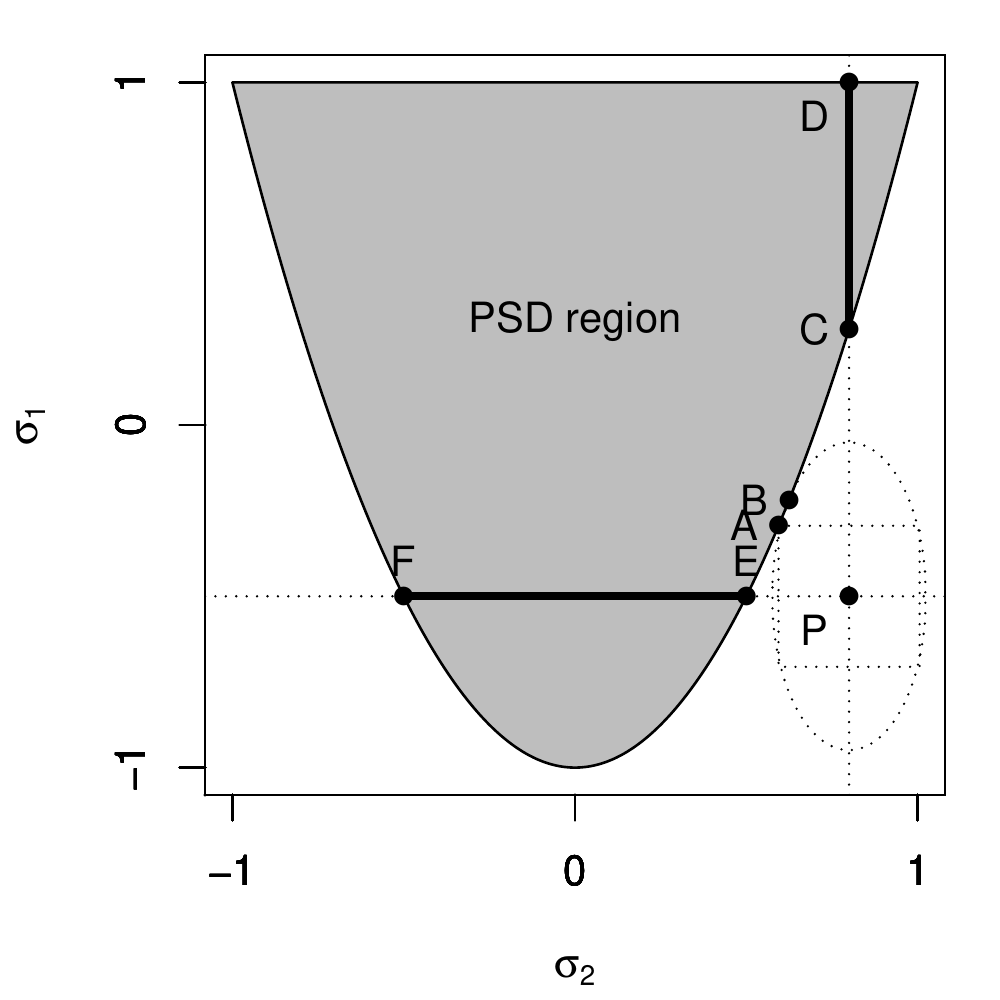}};
    \node at (8,1.8) {\includegraphics[scale=0.45]{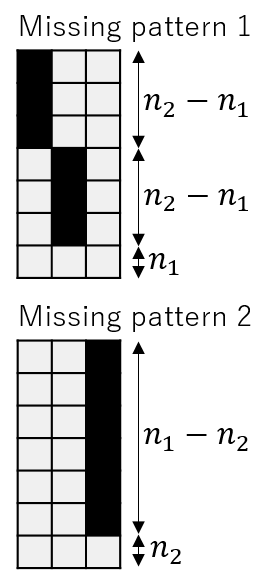}};
  \end{tikzpicture}
  \caption{(Left) Two-dimensional covariance matrix space.
  (Right) Two simple missing patterns. Black elements represent missing values.}
  % \end{center}
  \label{fig:example}
\end{figure}

The following simple example shows that our weighted formulation~\eqref{Sigmatilde} is better than the non-weighted formulation.
Consider three-dimensional data $X \in \mathbb{R}^{n\times 3}$.
To derive simple analytical results, we suppose that the pairwise covariance matrix and observation ratio matrix are
\begin{align*}
	\Spair =
	\begin{bmatrix}
		1 & s_1 & s_2\\
		s_1 & 1 & s_2\\
		s_2 & s_2 & 1
	\end{bmatrix}
  {\rm ~and~}
  R = \frac{1}{n}
	\begin{bmatrix}
		\cdot & n_1 & n_2\\
		n_1 & \cdot & n_2\\
		n_2 & n_2 & \cdot
	\end{bmatrix},
\end{align*}
respectively.
We restrict diagonal elements of the covariance estimate to $1$ for simplicity.
From symmetry of the problem, we can parameterize the covariance estimate as
$\Sigma = [ 1, \sigma_1, \sigma_2 ; \sigma_1, 1, \sigma_2 ; \sigma_2, \sigma_2, 1 ]$.
Here, we can see $\Spair$ and $\Sigma$ in the same two-dimensional space as in Figure~\ref{fig:example}.
Simple calculation yields that the PSD condition of $\Spair$ is $2s_2^2 -1 \leq s_1 \leq 1$,
shown in gray in Figure~\ref{fig:example}.
Hereafter, to show differences among the methods, without loss of generality we suppose $s_1 < 2s_2^2 - 1$ so that $\Spair$ is not PSD, and also $s_2\geq0$.
Point~P in Figure~\ref{fig:example} is an example of such an $\Spair$.

Consider the case where
$n_{1}$ is sufficiently small and $n_{2}$ is sufficiently large, which is realized by missing data pattern~1 in Figure~\ref{fig:example}.
The CoCoLasso (non-weighted max norm) solution is the tangent point of the gray region and an elliptic contour with center point~P, shown as point~A in Figure~\ref{fig:example}.
The solution of the non-weighted Frobenius norm is the tangent point of the gray region and a square contour with center point~P, shown as point~B.
The optimal point of HMLasso (weighted norm) will be close to point~C,
because weights $R_{13}$ and $R_{23}$ are much larger than $R_{12}$, so the optimal solution must be close to the tangent point of the gray PSD region and line~CD.
On the other hand, the sample covariance matrix $S$ with complete observations satisfies $\sigma_2 \approx s_2$ and $2\sigma_2^2 - 1 \leq \sigma_1 \leq 1$, represented as line segment~CD.
Since point~C is closer to any point on line segment~CD than are A or B,
the HMLasso estimate is always closer to the covariance matrix of the complete data than are estimates using non-weighted norms.

Consider another case where
$n_{1}$ is sufficiently large and $n_{2}$ is sufficiently small, which is realized by missing pattern~2 in Figure~\ref{fig:example}.
By reasoning similar to the case described above, the solutions for the non-weighted max and Frobenius norm are A and B, respectively, and the HMLasso optimal point will be close to point~E in Figure \ref{fig:example}.
The sample covariance matrix $S$ with complete observations satisfies $\sigma_1 \approx s_1$ and $2\sigma_2^2 - 1 \leq \sigma_1 \leq 1$, shown as line segment~EF.
Hence, the HMLasso estimate is again superior to those of the non-weighted norms.

\subsection{Algorithms}

\begin{algorithm}[t]
  \caption{Covariance Estimation with ADMM}
  \label{admm}
  \begin{algorithmic}
    \Require{$\Spair, W, \mu$}
    \For{$k = 1, 2, \dots$}
    \State $A_{k+1} \leftarrow $ the projection of $\left(B_k + \Spair + \mu \Lambda_k \right)$ \\
    ~~~~~~~~~~~~~~~~~~~~ onto the PSD space
    \State $B_{k+1} \leftarrow \left( A_{k+1} - \Spair - \mu\Lambda_k \right) \oslash \left( \mu W\odot W + I \right)$
    \State $\Lambda_{k+1} \leftarrow \Lambda_k - \frac{1}{\mu} \left( A_{k+1} - B_{k+1} - \Spair \right)$
    \EndFor
    \Ensure{$A_{i+1}$}
  \end{algorithmic}
\end{algorithm}

The Lasso optimization problem using the covariance matrix~\eqref{betahat} can be solved by various algorithms for the Lasso, such as the coordinate descent algorithm~\cite{Friedman:2010} and the least angle regression algorithm~\cite{Efron:2004}.
Our implementation uses the coordinate descent algorithm because it is efficient for high-dimensional data.
The algorithm details are described in Appendix \ref{apx:algorithms}.
We use the warm-start and safe screening techniques to speed up the algorithm.

We use the ADMM algorithm~\cite{Boyd:2011} to derive the PSD covariance matrix optimization~\eqref{Sigmatilde}, which can be rewritten as
\begin{align*}
  (A, B)
  &= \argmin_{A \succeq 0, B = A - \Spair} \| W \odot B \|_{\rm F}^2.
\end{align*}
Therefore, the augmented Lagrangian function is
\begin{align}
  f(A, B, \Lambda) = &\frac{1}{2} \| W \odot B \|_{\rm F}^2 - \langle \Lambda, A-B-\Spair \rangle \nonumber \\
  &+ \frac{1}{2\mu} \| A-B-\Spair\|_{\rm F}^2, \label{AugLagrangian}
\end{align}
where $\Lambda$ is a Lagrangian matrix and $\mu$ is an augmented Lagrangian parameter.
We iteratively update $A$, $B$, and $\Lambda$ subject to $A \succeq 0$ by minimizing~\eqref{AugLagrangian} in terms of each variable.
The resulting algorithm is similar to the CoCoLasso algorithm, except for the update rule for $B$ due to weight matrix $W$. To derive the $B$-step update equation, differentiating $f(A, B, \Lambda)$ with respect to $B$ yields
\begin{align*}
  \partial_B f(A, B, \Lambda) = W \odot W \odot B + \Lambda
  - \frac{1}{\mu} \left( A - B - \Spair \right).
\end{align*}
Solving $\partial_B f(A, B, \Lambda) = 0$, we obtain the update rule
\begin{align*}
  B \leftarrow (A - \Spair - \mu \Lambda) \oslash (\mu W \odot W + I).
\end{align*}
The algorithm for solving~\eqref{Sigmatilde} is presented as Algorithm~\ref{admm}.
The difference between the max norm and the Frobenius norm is trivial when we use ADMM.
Appendix \ref{apx:algorithms} also describes the algorithm for the weighted max norm.

\section{Theoretical Properties}

In this section, we investigate statistical properties of the proposed estimator.
We first obtain a refined non-asymptotic property for the pairwise covariance matrix, which explicitly includes missing rate effects.
We then derive a non-asymptotic property of our estimate in~\eqref{Sigmatilde}.
These results show that $\alpha=1$ weighting is superior in terms of non-asymptotic properties over other weighting ($\alpha \neq 1$) including non-weighted formulation ($\alpha=0$).
Note that we focus on the Frobenius norm formulation in~\eqref{Sigmatilde}, but we can see that $\alpha=1$ weighting is superior as well for the max norm formulation,
though CoCoLasso uses the non-weighted norm ($\alpha=0$).
Complete proofs for propositions and theorems in this section are given in Appendix \ref{apx:proofs}.

\subsection{Preliminaries}

Let $M=(M_{ij})\in\mathbb{R}^{n \times p}$ be the observation pattern matrix whose elements are 1 when data are observed and 0 otherwise, so that $Z = M \odot X$.
We suppose a sub-Gaussian assumption on $M$, which plays a key role in non-asymptotic properties.
This assumption often holds, as seen in the following proposition.

\begin{Definition}
	A random variable $X \in \mathbb{R}$ is said to be sub-Gaussian with $\tau^2$ if
    $
    	E\left[\exp(s(X-E[X]))\right] \leq \exp\left( {\tau^2s^2}/{2} \right)$ for all $s\in \mathbb{R}.
      $
    A random vector $X \in \mathbb{R}^p$ is said to be sub-Gaussian with $\tau^2$ if $v^\top X$ is a sub-Gaussian variable with $\tau^2$ for all $v \in \mathbb{R}^p$ satisfying $\| v\|_2=1$.
\end{Definition}
\begin{Assumption}
  \label{ass-subgaussian}
The rows of $M$ are independent and identically distributed with mean $\mu_M$, covariance $\Sigma_M$, and sub-Gaussian parameter $\tau^2$.
\end{Assumption}
\begin{Proposition}
  \label{prop-subgaussian}
	Assume that $M_{ij}$ values are independent and identically distributed as a Bernoulli distribution with mean $\mu_j$.
   Then, the rows of $M$ are sub-Gaussian with $\tau^2 = \max_{j} \mu_j(1-\mu_j) \leq 1/4$.
\end{Proposition}

For the theoretical analysis in this section, we substitute
$
\Sigmahat := \frac{1}{n}Z^\top Z \oslash \Pi
$
for $\Spair$, where $\Pi := (\pi_{jk}) := \Sigma_M + \mu_M\mu_M^\top$.
This is reasonable because $\Spair = \frac{1}{n}Z^\top Z \oslash R$ from~\eqref{Simppair}, and
the expectation for $R$ is ${\rm E}[R]= {\rm E}[\frac{1}{n}M^\top M] = \Pi
$.
We respectively call $\pi_{jk}$ and $1-\pi_{jk}$ the observed and missing rate, since they are expectations for the observed and missing ratios.
We suppose that $\mu_M$ and $\Sigma_M$ are known.
This substitution and assumption were also used in previous theoretical research~\cite{Loh:2012,Datta:2017}.

\subsection{Statistical Properties}

We first derive a refined non-asymptotic property of $\Sigmahat$.

\begin{Theorem}
  \label{TheoremSigmahat}
  Under Assumption~\ref{ass-subgaussian}, we have, for all $\varepsilon \leq c\tau^2 X_{\max}^2/\pi_{jk}$,
\begin{align*}
\Pr \left( \left| \Sigmahat_{jk} - S_{jk} \right| \leq \varepsilon \right)
\geq 1 - C \exp \left( - cn \varepsilon^{2} \pi_{jk}^2\zeta^{-1} \right),
\end{align*}
where $\zeta = \tau^2 X_{\max}^4 \max \left\{ \tau^2, \mu_j^2, \mu_k^2 \right\}$, $X_{\max} = \max_{i,j} |X_{ij}|$, and $C$ and $c$ are universal constants.
\end{Theorem}

\begin{sproof}
  We see that
  \begin{align*}
    &\left| \Sigmahat_{jk} - S_{jk} \right|
    \leq \frac{1}{n \pi_{jk}} \left| \sum_{i=1}^{n} v_{ijk} (m_{ij} - \mu_j)(m_{ik} - \mu_k) \right| \nonumber\\
    & + \frac{\mu_j}{n \pi_{jk}} \left| \sum_{i=1}^{n} v_{ijk} (m_{ik} - \mu_k) \right|
     + \frac{\mu_k}{n \pi_{jk}} \left| \sum_{i=1}^{n} v_{ijk} (m_{ij} - \mu_j) \right|,
  \end{align*}
  with $v_{ijk} := x_{ij}x_{ik}$.
  The first term is bounded
  using Lemma B.1 in~\cite{Datta:2017},
  and the second and third terms are bounded
  using Property~(B.2) in~\cite{Datta:2017}.
  Careful analyses considering $\pi_{jk}, \mu_j$, and $ \mu_k$ yield the assertion.
\end{sproof}

In Theorem~\ref{TheoremSigmahat}, the missing rate appears explicitly and the non-asymptotic property is stricter than Definition~1 and Lemma~2 in~\cite{Datta:2017}.
To clearly see the missing rate effect, we replace $\varepsilon$ by $\varepsilon/\pi_{jk}$.
Then, for all $\varepsilon \leq c\tau^2 X_{\max}^2$ we have
\begin{align*}
\Pr \left( \pi_{jk} \left| \Sigmahat_{jk} - S_{jk} \right| \leq \varepsilon \right)
\geq 1 - C \exp \left( - cn \varepsilon^{2}\zeta^{-1} \right).
\end{align*}
Since the right side does not depend on $\pi_{jk}$, we can see that the concentration probability of $\pi_{jk} | \Sigmahat_{jk} - S_{jk}|$ is equally bounded regardless of the missing rate.
This implies that our weighted formulation balances uncertainty of each element of $\Sigmahat$,
while non-weighted formulations such as CoCoLasso suffer from this imbalance.

Next, we derive a non-asymptotic property of our weighted estimator $\Sigmatilde$.

\begin{Theorem}
\label{TheoremSigmatilde}
Under Assumption~\ref{ass-subgaussian},
we have, for all $\varepsilon \leq c\tau^2X_{\max}^2 (\min_{j, k} W_{jk}/\pi_{jk}) / W_{\min}$,
\begin{align*}
	&\Pr \left( \frac{1}{p^2} \left\| \Sigmatilde - S \right\|_{\rm F}^2 \leq \varepsilon^2 \right) \nonumber \\
    &\geq 1 - p^2 C \exp \left( -cn \varepsilon^2 W_{\min}^2 \left(\min_{j, k}\frac{\pi_{jk}}{W_{jk}}\right)^2 \zeta^{-1}\right),
\end{align*}
where $\zeta = \tau^2 X_{\max}^4 \max \left\{ \tau^2, \mu_1^2,\dots, \mu_p^2 \right\}$, $W_{\min} = \min_{j,k} W_{jk}$, $X_{\max} = \max_{i,j} |X_{ij}|$, and $C$ and $c$ are universal constants.
\end{Theorem}

\begin{sproof}
  We have
  $
  \| W \odot (\Sigmatilde - S)\|_{\rm F}
    \leq \| W \odot (\Sigmatilde - \Sigmahat)\|_{\rm F} + \| W \odot (\Sigmahat - S)\|_{\rm F}
    \leq 2\| W \odot (\Sigmahat - S)\|_{\rm F}
  $
  by the triangular equation and the optimality of $\Sigmatilde$.
  Using
  $
    W_{\min}^2 \| \Sigmatilde - S \|_{\rm F}^2
  	\leq 	\| W \odot ( \Sigmatilde - S ) \|_{\rm F}^2.
  $
  and Theorem~\ref{TheoremSigmahat} yields the assertion.
\end{sproof}

According to Theorem~\ref{TheoremSigmatilde}, we can see that the weighted Frobenius norm with $\alpha=1$ is superior to the other weightings.
Let $\pi_{\min} = \min_{j,k} \pi_{jk}$ and $\pi_{\max} = \max_{j,k} \pi_{jk}$.
Substituting $W_{jk}=\pi_{jk}^\alpha$, we have
the concentration probability
 \begin{align*}
   1 - p^2 C \exp \left( -cn \varepsilon^2 \pi_{\min}^{2\alpha} \left(\min_{j, k} \pi_{jk}^{2-2\alpha}\right) \zeta^{-1}\right),
\end{align*}
with the constraint $\varepsilon \leq c\tau^2X_{\max}^2 ( \min_{j, k} \pi_{jk}^{\alpha-1} ) / \pi_{\min}^\alpha$. \
We divide this into two cases: $0 \leq \alpha \leq 1$ and $\alpha \geq 1$.
i)~For $0 \leq \alpha \leq 1$, the concentration probability becomes
$1 - p^2 C \exp \left( -cn \varepsilon^2 \pi_{\min}^{2} \zeta^{-1}\right),$
and the constraint of $\varepsilon$ becomes
$\varepsilon \leq c\tau^2X_{\max}^2 \left(\pi_{\max}/\pi_{\min}\right)^\alpha / \pi_{\max}$.
Since $\pi_{\max}/\pi_{\min} \ge 1$, the constraint region of $\varepsilon$ is maximized at $\alpha=1$. \
ii)~For $\alpha \geq 1$, the concentration probability becomes
$1 - p^2 C \exp \left( -cn \varepsilon^2 \pi_{\max}^{2} (\pi_{\min}/\pi_{\max})^{2\alpha} \right)$
and the constraint of $\varepsilon$ becomes
$\varepsilon \leq c\tau^2X_{\max}^2 / \pi_{\min}$.
Since $\pi_{\min}/\pi_{\max} \le 1$, the concentration probability is maximized at $\alpha=1$. \
These imply that the case where $\alpha=1$ is superior to the other weighted norm formulation, because the non-asymptotic bound for $\alpha=1$ holds with the highest probability under the loosest condition.

\section{Numerical Experiments}

We conducted some experiments using both synthetic and real-world data.
More comprehensive simulation results under various conditions were given in Appendix \ref{apx:experiments}.

\subsection{Simulations with Various Norms}

First, we investigated the effect of various weighted norms in~\eqref{Sigmatilde}.
We compared the max norm with $\alpha=0, 1/2, 1, 2$ and the Frobenius norm with $\alpha=0, 1/2, 1, 2$.
The max norm with $\alpha=0$ corresponds to CoCoLasso, and the Frobenius norm with $\alpha=0$ corresponds to a simple projection
onto the PSD region.

Training data were $X \in \mathbb{R}^{n \times p}$ with $n=10000$ and $p=100$ generated by $ \mathcal{N}(0, \Sigmastar)$ with $\Sigmastar_{jk}=0.5$ for $j \neq k$ and $\Sigmastar_{jk}=1$ for $j=k$.
Responses $y$ were defined as $y = X \beta + \varepsilon$ with $\beta_1 = 10, \beta_{11} = -9, \beta_{21} = 8, \beta_{31} = -7, \dots, \beta_{91} = -1$, and $\beta_j = 0$ otherwise, and $\varepsilon \sim \mathcal{N}(0, 1)$.
We introduced missing values completely at random, setting a missing rate for each column sampled from a uniform distribution $U(0,1)$.
Test data were generated independently in the same manner, except that
we did not introduce missing values for evaluation.

The regularization parameter $\lambda$ was selected by calibrated five-fold cross validation~\cite{Datta:2017}.
We iterated each experiment 30 times and plotted averaged results with standard errors.

\begin{figure}[t]
  % \begin{center}
  \centering
    \includegraphics[height=4cm]{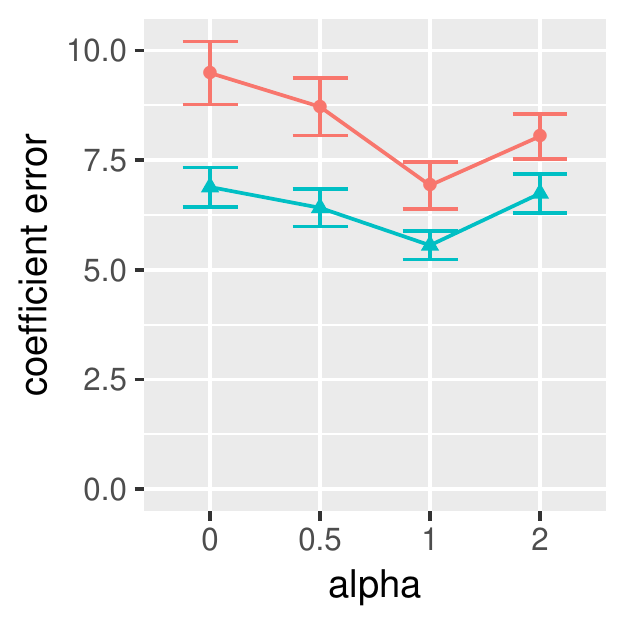}
    \includegraphics[height=4cm, width=4.8cm]{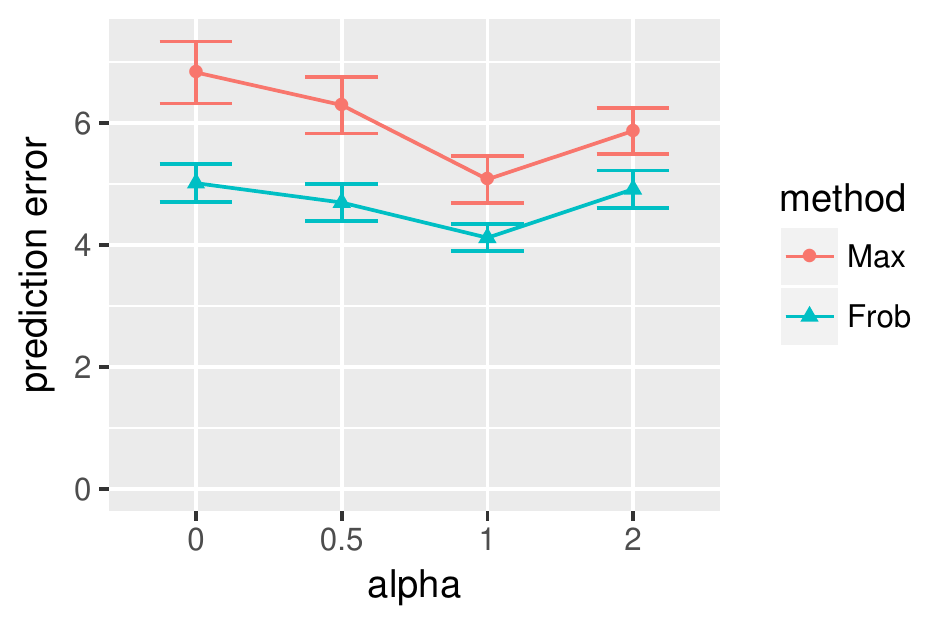}
  \caption{Comparison of the max and Frobenius norms with \\ $\alpha=0, 0.5, 1, 2$.}
  % \end{center}
  \label{fig:9comparison}
\end{figure}

Figure~\ref{fig:9comparison} shows the results.
The performance measures were
the $\ell_2$ error for the regression coefficients,
and the root mean square error of prediction.
The weighted norms with $\alpha=1$ were effective for both the Frobenius and max norm formulations, as suggested by the non-asymptotic theoretical analyses.
In addition, the Frobenius norms outperformed the max norms.
Therefore, we use the Frobenius norm with $\alpha=1$ as the proposed method, namely, HMLasso, in subsequent experiments.
In contrast, CoCoLasso (the max norm with $\alpha=0$) was apparently inferior to HMLasso.

\subsection{Simulations Under Various Conditions}

\begin{figure}[t]
  \centering

  \begin{tikzpicture}
    % \draw [help lines] (0,0) grid (6,6);
    \node at (2,5) {\includegraphics[height=4cm]{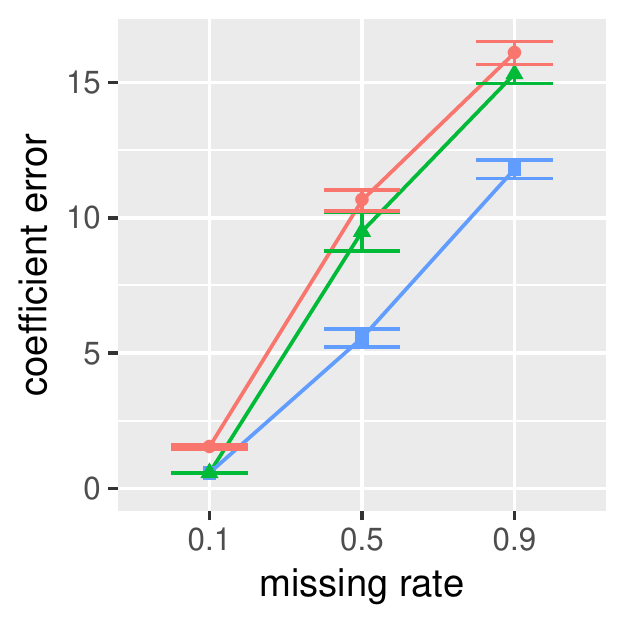}};
    \node at (6.5,5) {\includegraphics[height=4cm]{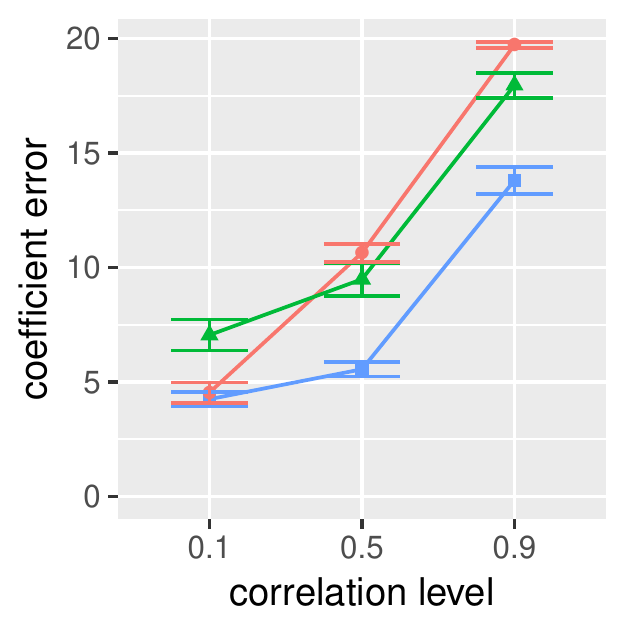}};
    \node at (2,1) {\includegraphics[height=4cm]{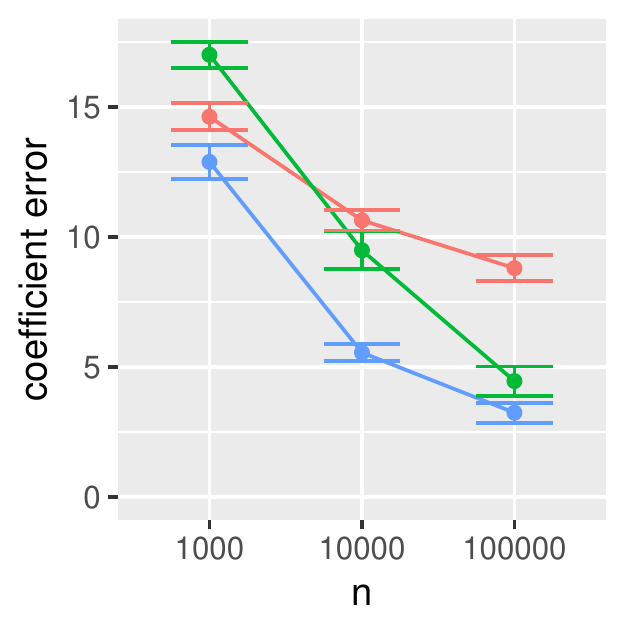}};
    \node at (7.5,1) {\includegraphics[height=4cm]{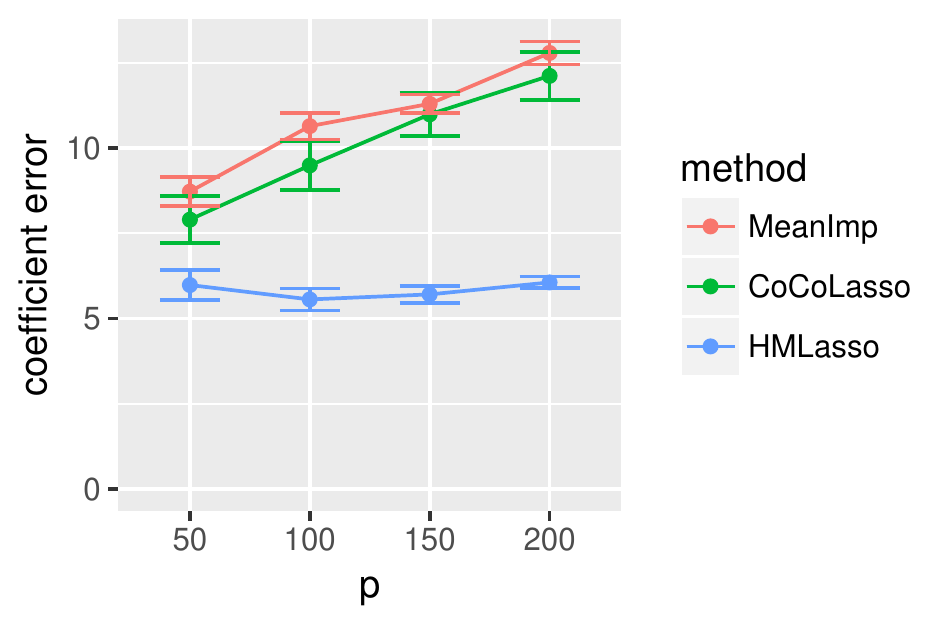}};
  \end{tikzpicture}
    \caption{Simulation results with various missing rates, covariance levels, sample sizes, and dimension numbers.
    }
   \label{fig:SimConditions}
\end{figure}

We next compared the performance of HMLasso with other methods under various conditions,
examining the mean imputation method,
CoCoLasso (the max norm with $\alpha=0$), and HMLasso (the Frobenius norm with $\alpha=1$).
Following the simulation setting in the previous subsection, we varied the missing data rate, covariance level, sample size, and number of variables.
We set the missing rates to $\mu = 0.1, 0.5, 0.9$,
the covariance levels to $r = 0.1, 0.5, 0.9$,
the sample size to $n=10^3, 10^4, 10^5$,
and the number of variables to $p=50, 100, 150, 200$.
Note that we also examined other missing imputation methods such as {\tt mice}~\cite{Buuren:2011} and {\tt missForest}~\cite{Stekhoven:2012}, but their computational costs were over $100$ times larger than those for the above methods, so we excluded these methods in our experiments.

Figure~\ref{fig:SimConditions} shows the results.
HMLasso outperformed other methods under almost all conditions,
especially for data with high missing rates and high covariance levels.
In addition, high dimensionality did not adversely affect HMLasso,
while the other methods showed gradually worse performance.

\subsection{Residential Building Dataset}

\begin{figure}[t]
  \centering
  \includegraphics[height=4cm]{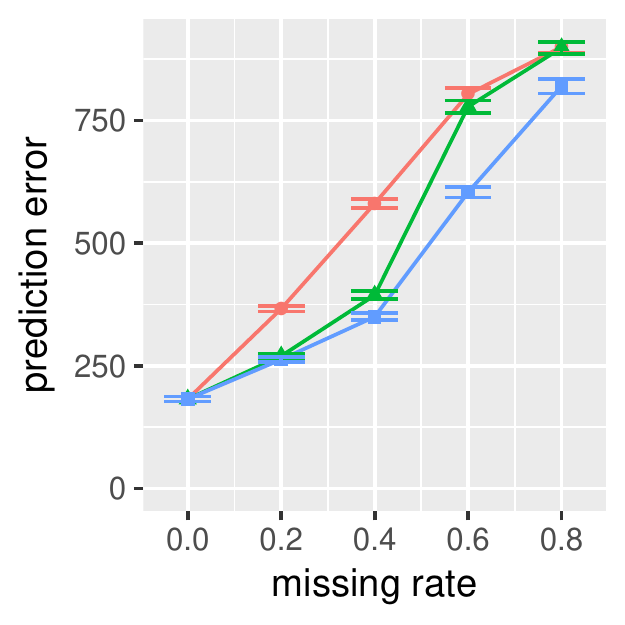}
  \includegraphics[height=4cm]{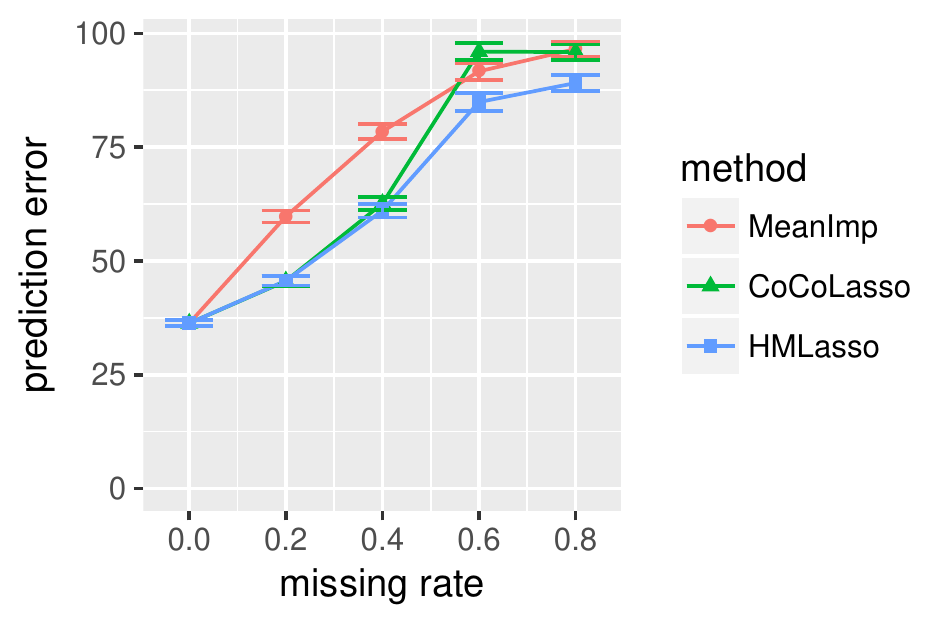}
    \caption{Analysis of residential building data with various patterns and rates for missing data.
    The outcomes are sales prices (left) and construction costs (right).}
    \label{fig:residential}
\end{figure}

We evaluated the performance using a real-world residential building dataset~\cite{Rafiei:2015} from the UCI datasets repository\footnote{https://archive.ics.uci.edu/ml/datasets/Residential+Building+Data+Set}.
The dataset included construction costs, sale prices, project variables, and economic variables corresponding to single-family residential apartments in Tehran.
The objective was to predict sale prices and construction costs from physical, financial, and economic variables.
The data consisted of $n=372$ samples and $p=105$ variables, including two output variables.
We introduced missing values at missing rates $\mu = 0, 0.2, 0.4, 0.6, 0.8$.
Since we cannot obtain the true coefficient, we evaluated performance in terms of prediction error from complete samples.
We randomly split data into 300 samples for training, 36 for validation, and 36 for testing, and iterated the experiments 30 times.

Figure~\ref{fig:residential} shows the results.
HMLasso outperformed the other methods under nearly all cases.
The advantage of HMLasso was clear especially for high missing rates.

\section{Conclusion}
We proposed a novel regression method for high-dimensional data with high missing rates.
We demonstrated the advantages of our method through theoretical analyses and numerical experiments.
The proposed method allows utilization of previously unusable data with high missing rates.

\newpage

\bibliographystyle{plain}
\bibliography{ijcai_2019_hmlasso}

\newpage
% \onecolumn
\renewcommand{\thesection}{\Alph{section}}
\setcounter{section}{0}

\section*{Appendix}

\section{Discussions from Asymptotic Perspective}
\label{apx:asymptotic}
We describe another view of our weighted formulation.
Note that this is a rough result, but we intuitively interpret why weighted norm with $\alpha=1/2$ performs well from the asymptotic perspective ($n \gg p$).

The standard asymptotic theory shows that we have, for a large pairwise observation number $n_{jk}$,
\begin{align*}
	\sqrt{n_{jk}} (\Spair_{jk} - \Sigmastar_{jk}) \sim \mathcal{N}(0, \tau_{jk}^2),
\end{align*}
where $\Sigmastar$ is a population covariance matrix and $\tau_{jk}$ is a constant.
Here we assume that $\Spair_{jk}$'s are independent and $\tau_{jk}=\tau$ for all $j, k$.
% Although this assumption does not hold in general,
Then, the likelihood of $\Spair$ can be approximated to
\begin{align*}
\prod_{j,k} \frac{1}{\sqrt{2\pi\tau^2}} \exp \left( - \frac{1}{2\tau^2} \left( \sqrt{n_{jk}}(\Spair_{jk} - \Sigmastar_{jk})\right)^2 \right).
\end{align*}
Hence, the maximum likelihood estimator of $\Sigmastar$ under the PSD constraint can be approximated to
\begin{align*}
\argmin_{\Sigma \geq 0} \sum_{j,k} n_{jk} (\Spair_{jk} - \Sigma_{jk})^2,
\end{align*}
which is equivalent to our method with $\alpha={1}/{2}$.

\section{Algorithms}
\label{apx:algorithms}

\subsection{Cordinate Descent Algorithm with the Covariance Matrix}
We describe an algorithm for solving the Lasso problem using the covariance matrix (\ref{betahat}).
Let $\mathscr{L}(\beta)$ be the objective function of \eqref{betahat}.
To derive the update equation, when $\beta_j \neq 0$, differentiating $\mathscr{L}(\beta)$ with respect to $\beta_j$ yields
\begin{align*}
\partial_{\beta_j} \mathscr{L}(\beta) =&
\Sigmatilde_{j, -j} \beta_{-j} + \Sigmatilde_{jj} \beta_j
- \rhopair_j + \lambda \sgn(\beta_j),
\end{align*}
where $\beta_{-j}$ denotes $\beta$ without the $j$-th component, and $X_{j,-j}$ denotes the $j$-th row of $X$ without the $j$-th column.
Solving $\partial_{\beta_j} \mathscr{L}(\beta) = 0$, we obtain the update rule as
\begin{align*}
\beta_j \leftarrow \frac{1}{\Sigmatilde_{jj}}
S\left( \left(\rhopair_j - \Sigmatilde_{j, -j}\beta_{-j} \right), \lambda \right), \label{eq:cdaupdate}
\end{align*}
where $S(z, \gamma)$ is a soft thresholding function
\begin{align*}
S(z, \gamma) :&= {\rm sgn}(z)(|z|-\gamma)_+ \\
&= \begin{cases}
z - \gamma & {\rm if} \ z>0 \ {\rm and} \ \gamma < |z|,\\
z + \gamma & {\rm if} \ z<0 \ {\rm and} \ \gamma < |z|,\\
0 & {\rm if} \ |z| \leq \gamma.
\end{cases}
\end{align*}
The whole algorithm for the Lasso-type optimization problem \eqref{betahat} is described in Algorithm \ref{cda}.

\begin{algorithm}[t]
  \caption{Lasso with Covariance Matrix}
  \label{cda}
  \begin{algorithmic}
    \Require{$\Sigmatilde$, $\rhopair$, $\lambda$}
    % \Function{covlasso}{$\Sigma, \rho$}
		\State initialize $\beta$
		\While{until convergence}
		\For{$j = 1, \cdots, p$}
		\State $\beta_j \leftarrow \frac{1}{\Sigmatilde_{jj}}
		S \left( \left( \rhopair_j - \Sigmatilde_{j, -j} \beta_{-j}\right), \lambda \right) $
    \EndFor
		\EndWhile
    % \EndFunction
    \Ensure{$\beta$}
  \end{algorithmic}
\end{algorithm}

\subsection{ADMM for the Weighted Max Norm Formulation}

We describe the ADMM algorithm for the weighted max norm formulation.
This is a natural extension of the CoCoLasso algorithm.
The difference from the CoCoLasso algorithm is B-step update in ADMM described in Algorithm \ref{admm-max}.

\begin{algorithm}[t]
  \caption{B-step Update for max norm in ADMM}
	\label{admm-max}
  \begin{algorithmic}
    \Require{$A_{k+1}, \Lambda_k, \Sigmahat, \mu, W$}
    \State define $c=\vecl\left(A_{k+1} - \Sigmahat - \mu\Lambda_k\right)$, $w=\vecl(W)$
    \State sort $c$ as $w_1|c_1| \geq w_2|c_2| \geq \dots$
    \State find $l = \max_{l'} \left\{ l' : w_l'|c_l'| - \frac{\left( \sum_{j=1}^{l'} |c_j| \right) - \frac{\mu}{2}}{\sum_{j=1}^{l'} \frac{1}{w_j}} >0 \right\}$
    \State define $d = \frac{\left( \sum_{j=1}^{l} |c_j| \right) - \frac{\mu}{2}}{\sum_{j=1}^{l} \frac{1}{w_j}}$
    \State define $B_{k+1}=\matl(b)$ such that $b_j=c_j$ for $|c_j| \leq \frac{d}{w_j}$, and $b_j = \frac{d \sgn(c_j)}{w_j}$ for $|c_j| > \frac{d}{w_j}$
    \Ensure{$B_{k+1}$}
  \end{algorithmic}
\end{algorithm}

\section{Proofs}
\label{apx:proofs}
\subsection{Proof of Proposition \ref{prop-subgaussian}}
\begin{proof}
	First, we prove that random variables with Bernoulli distribution is sub-Gaussian.
	Because $m_{ij} \sim Bernoulli(\mu_j)$, we have $E[m_{ij}] = \mu_j$ and
    \begin{align*}
    	&E[\exp(s(m_{ij}-\mu_j))]\nonumber \\
       =& \mu_j \exp(s(1-\mu_j)) + (1-\mu_j) \exp(-s\mu_j).
    \end{align*}
    Hence, the Taylor expansion yields
    \begin{align*}
       &\log E[\exp(s(m_{ij}-\mu_j))] \\
       =& -s\mu_j + \log (1 + \mu_j (\exp(s)-1))\nonumber\\
       \leq& \frac{\mu_j(1-\mu_j) s^2}{2},
    \end{align*}
    which indicates $m_{ij}$ is sub-Gaussian with $\tau_j^2 = \mu_j(1-\mu_j)$.
   We can see $\tau_j^2 \leq 1/4$ since $\mu_j \in [0,1]$.

	 Next, we prove the proposition.
	 For a random vector $M_i$, the $i$-th row of $M$, we have
	 \begin{align*}
		 &E[\exp (s (v^\top M_i - E[v^\top M_i]))]\\
		 =& E[\exp (s (v^\top M_i - E[v^\top \mu]))]\\
		 =& \prod_{j=1}^p E[\exp (s v_j (M_{ij} - \mu_j))]\\
		 \leq& \prod_{j=1}^p \exp\left( \frac{\tau_j^2 v_j^2 s^2}{2}\right)\\
		 =& \exp\left( \frac{\left(\sum_{j=1}^p \tau_j^2 v_j^2 \right) s^2}{2}\right)\\
		 \leq& \exp\left( \frac{s^2 \max_j \tau_j^2}{2}\right),
	 \end{align*}
	 for any unit vector $v$.
\end{proof}

\subsection{Proof of Theorem \ref{TheoremSigmahat}}

\begin{proof}
  We see that
  \begin{align*}
    \left| \Sigmahat_{jk} - S_{jk} \right|
    &= \left| \frac{1}{n} \sum_{i=1}^{n} m_{ij}m_{ik} x_{ij}x_{ik} / \pi_{jk} - \frac{1}{n} \sum_{i=1}^{n} x_{ij}x_{ik} \right|\\
    &\leq \frac{1}{\pi_{jk}} \frac{1}{n} \left| \sum_{i=1}^{n} x_{ij}x_{ik} (m_{ij} - \mu_j)(m_{ik} - \mu_k) \right| \nonumber\\
    & + \frac{\mu_j}{\pi_{jk}} \frac{1}{n} \left| \sum_{i=1}^{n} x_{ij}x_{ik} (m_{ik} - \mu_k) \right| \nonumber \\
    & + \frac{\mu_k}{\pi_{jk}} \frac{1}{n} \left| \sum_{i=1}^{n} x_{ij}x_{ik} (m_{ij} - \mu_j) \right|.
  \end{align*}
  We denote the three terms on the right-hand side by $T_1, T_2,$ and $T_3$, respectively.

  (T1):
  Let $v_i:=x_{ij}x_{ik}$.
  Then we have $\|v\|_\infty \leq X_{\max}^2$.
  Remember that $m_{ij} - \mu_j$ and $m_{ik} - \mu_k$ are sub-Gaussian with parameter $\tau^2$.
  Then, by applying Lemma B.1 in CoCoLasso, we have
  \begin{align*}
    &\Pr(T_1 > \varepsilon)\\
    =& \Pr \left(\frac{1}{n} \left| \sum_{i=1}^{n} x_{ij}x_{ik} (m_{ij} - \mu_j)(m_{ik} - \mu_k) \right| > \pi_{jk} \varepsilon \right)\\
    \leq& C \exp \left( - \frac{cn \varepsilon \pi_{jk}^2}{\tau^4 X_{\max}^4} \right)
  \end{align*}
  for all $\pi_{jk} \varepsilon \leq c\tau^2 X_{\max}^2$, i.e., $\varepsilon \leq c\tau^2 X_{\max}^2 / \pi_{jk}$.

  (T2) and (T3):
  By property (B.2) in CoCoLasso, we can see that for any vector $v$ and independent sub-Gaussian vector $w_i$ with parameter $\tau^2$, we have
  \begin{align*}
    \Pr \left( \frac{1}{n} \left| \sum_{i=1}^n v_i w_i \right| > \varepsilon \right)
    \leq C \exp \left( - \frac{cn^2\varepsilon^2}{\|v\|_2^2\tau^2} \right).
  \end{align*}
  If we define $v_i:=x_{ij}x_{ik}$, we have $\|v\|_2^2 \leq nX_{\max}^4$.
  Remember that $m_{ij} - \mu_j$ and $m_{ik} - \mu_k$ are sub-Gaussian with parameter $\tau^2$.
  Hence, we have
  \begin{align*}
    &\Pr(T_2>\varepsilon)\\
    =& \Pr \left( \frac{1}{n} \left| \sum_{i=1}^{n} x_{ij}x_{ik} (m_{ik} - \mu_k) \right| > \frac{\pi_{jk}\varepsilon}{\mu_j} \right)\\
    \leq& C \exp \left( - \frac{cn \varepsilon^2 \pi_{jk}^2}{\tau^2X_{\max}^4\mu_j^2} \right).
  \end{align*}
  Similarly, we have
  \begin{align*}
    &\Pr(T_3>\varepsilon)\\
    =& \Pr \left( \frac{1}{n} \left| \sum_{i=1}^{n} x_{ij}x_{ik} (m_{ij} - \mu_j) \right| > \frac{\pi_{jk}\varepsilon}{\mu_k} \right)\\
    \leq& C \exp \left( - \frac{cn \varepsilon^2 \pi_{jk}^2}{\tau^2X_{\max}^4\mu_k^2} \right).
  \end{align*}

  Putting all together, we obtain that for all $\varepsilon \leq c\tau^2 X_{\max}^2 / \pi_{jk}$,
  \begin{align*}
    &\Pr \left( \left| \Sigmahat_{jk} - S_{jk} \right| > \varepsilon \right)\\
%     &\leq \sum_{j=1}^3 \Pr \left( T_j > \varepsilon \right) \\
    \leq& C \exp \left(- \frac{cn \varepsilon^2 \pi_{jk}^2}{\tau^2 X_{\max}^4 \max \{ \tau^2, \mu_j^2, \mu_k^2 \}} \right).
  \end{align*}
\end{proof}

\subsection{Proof of Theorem \ref{TheoremSigmatilde}}

\begin{proof}
Since $\Sigmatilde = \argmin_{\Sigma \succeq 0} \| W \odot (\Sigma - \Sigmahat)\|_{\rm F}^2$,
we have, using the triangular equation,
\begin{align*}
  &\| W \odot (\Sigmatilde - S)\|_{\rm F}\\
  \leq& \| W \odot (\Sigmatilde - \Sigmahat)\|_{\rm F} + \| W \odot (\Sigmahat - S)\|_{\rm F}\\
  \leq& 2\| W \odot (\Sigmahat - S)\|_{\rm F}.
\end{align*}
From Theorem \ref{TheoremSigmahat}, we have
\begin{align*}
	&\Pr \left( \| W \odot (\Sigmahat - S)\|_{\rm F} > \varepsilon \right)\\
    =& \Pr \left( \sum_{j,k} W_{jk}^2 \left( \Sigmahat_{jk} - S_{jk}\right)^2 > \varepsilon^2 \right)\\
    \leq& \sum_{j,k} \Pr \left(W_{jk}^2 \left( \Sigmahat_{jk} - S_{jk}\right)^2 > \varepsilon^2 p^{-2} \right)\\
    \leq& p^2 \max_{j, k} \Pr \left(W_{jk} \left| \Sigmahat_{jk} - S_{jk}\right| > \varepsilon p^{-1} \right)\\
    \leq& p^2 C \exp \left( -cn\varepsilon^2 p^{-2} \left( \min_{j, k} \frac{\pi_{jk}}{w_{jk}} \right)^2 \zeta^{-1}\right),
\end{align*}
for all $\varepsilon \leq cp\tau^2X_{\max}^2 \min_{j, k} \left(w_{jk}/\pi_{jk}\right)$,
where $\zeta = \max \left\{ \tau^2, \mu_1^2, \dots, \mu_p^2 \right\}$.
Hence, we have
\begin{align*}
	&\Pr \left( \| W \odot (\Sigmatilde - S)\|_{\rm F} > \varepsilon \right)\\
    \leq& \Pr \left( \| W \odot (\Sigmahat - S)\|_{\rm F} > \varepsilon/2 \right)\\
    \leq& p^2 C \exp \left( -cn\varepsilon^2 p^{-2} \left( \min_{j, k} \frac{\pi_{jk}}{w_{jk}} \right)^2 \zeta^{-1}\right),
\end{align*}
for all $\varepsilon \leq cp\tau^2X_{\max}^2 \min_{j, k} \left(\frac{w_{jk}}{\pi_{jk}}\right)$.
This is equivalent to
\begin{align*}
	&\Pr \left( \frac{1}{p^2} \left\| W \odot (\Sigmatilde - S)\right\|_{\rm F}^2 > \varepsilon^2 \right)\\
    \leq& p^2 C \exp \left( -cn \varepsilon^2 \left(\min_{j, k}\frac{\pi_{jk}}{W_{jk}}\right)^2 \zeta^{-1}\right).
\end{align*}
Using the inequality
$
W_{\min}^2 \left\| \Sigmatilde - S \right\|_{\rm F}^2
	\leq 	\left\| W \odot \left( \Sigmatilde - S \right) \right\|_{\rm F}^2,
$
we have
\begin{align*}
	&\Pr \left( \frac{1}{p^2} \left\| \Sigmatilde - S \right\|_{\rm F}^2 > \varepsilon^2 \right)\\
    \leq& p^2 C \exp \left( -cn \varepsilon^2 W_{\min}^2 \left(\min_{j, k}\frac{\pi_{jk}}{W_{jk}}\right)^2 \zeta^{-1}\right),
\end{align*}
for $\varepsilon \leq c\tau^2X_{\max}^2 (\min_{j, k} W_{jk}/\pi_{jk}) / W_{\min}$.
\end{proof}

\section{Numerical Experiments}
\label{apx:experiments}

We discribe additional simulation results.
The conditions include various missing patterns, covariance patterns, noise levels, and true parameters.

\subsection{Missing Patterns and Missing Rates}

We examined three missing patterns and three missing rates, resulting in nine conditions.
We set the missing rates to $\mu = 0.1, 0.5, 0.9$.
We introduced missing values according to the following missing patterns, which are thought to be common in real-world data.
\begin{enumerate}[(1)]
	\item Random pattern:
	Missing elements were selected with the same probability for all the elements.
	\item Column pattern:
	Missing rates differ for each column.
	The $j$-th column missing rate $\mu_j$ was sampled from the uniform distribution so that the overall missing rate was $\mu$.
	$\mu_j$ was sampled from $U(0, 0.2)$ for $\mu = 0.1$, from $U(0, 1)$ for $\mu = 0.5$, and from $U(0.8, 1)$ for $\mu = 0.9$.
	\item Row column pattern:
	Missing rates differ for each row and each column.
	The $(i, j)$-th element missing rate $\mu_{ij}$ was set so that the overall missing rate was $\mu$.
	Specifically, we defined $\mu_{ij} = \mu^i \mu_j$ where $\mu^i$ and $\mu_j$ were sampled from $U(0, 0.632)$ for $\mu=0.1$,
	$\mu_{ij} = \mu^i \mu_j$ where $\mu^i$ and $\mu_j$ were sampled from $U(0.414, 1)$ for $\mu=0.5$, and
	$\mu_{ij} = 1 - (1 - \mu^i) (1 - \mu_j)$ where $\mu^i$ and $\mu_j$ were sampled from $U(0.368, 1)$ for $\mu=0.9$.
\end{enumerate}

\begin{figure}
	\centering
	\includegraphics[height=3.3cm]{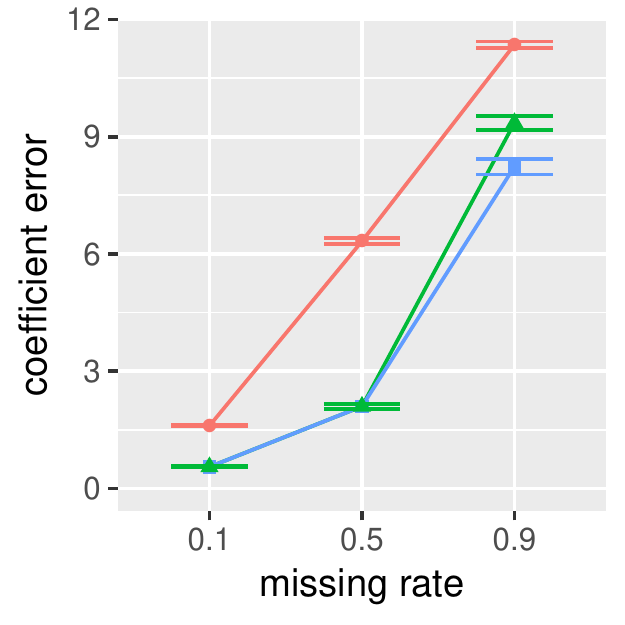}
	\includegraphics[height=3.3cm]{figures/ijcai_coef_miss_method.pdf}
	\includegraphics[height=3.3cm]{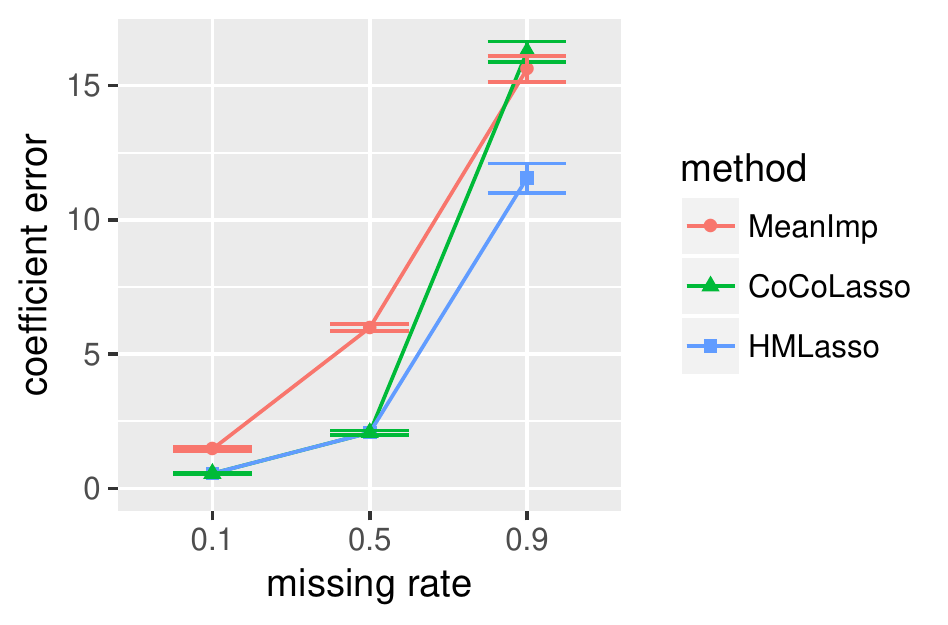}
	\caption{Simulation results with various missing patterns: random pattern (left), column pattern (center), and row column pattern (right).}
	\label{fig:miss_pt}
\end{figure}

Figure~\ref{fig:miss_pt} shows the results.
HMLasso outperformed other methods, when the missing rate was moderate or high.
In particular, in the cases of the column pattern and row column pattern, HMLasso delivered significant improvements.
This might be because the number of pairwise observations were very small for these missing patterns.
The mean imputation and CoCoLasso suffered from highly missing variables, while HMLasso suppressed the effects of them.

Note that the column and row missing patterns often appear in practice.
The column missing pattern appears when some variables are frequently observed and others are rarely observed.
This is typically caused by different data collection cost for each variable.
The row missing pattern appears when some samples are filled and other samples are highly missing.
This happens when some samples are considered to be important and they are frequently measured.

\subsection{Covariance Patterns and Covariance Levels}

We examined three covariance patterns and three covariance levels, resulting in nine conditions.
We set the covariance levels to $r = 0.1, 0.5, 0.9$.
The covariance matrix was generated according to the following covariance matrix patterns.
\begin{enumerate}[(1)]
	\item Uniform pattern:
	Covariances were uniform among all variables, where $\Sigmastar_{jk}=r$ for $j \neq k$ and $\Sigmastar_{jk}=1$ for $j=k$.
	\item Autoregressive pattern:
	Covariances among neighbors were strong, such that $\Sigmastar_{jk}=r^{|j - k|}$ for $j \neq k$ and $\Sigmastar_{jk}=1$ for $j=k$.
	\item Block pattern:
	All of the variables were divided into some blocks.
	The intra-block covariances were strong and inter-block covariances are zeros.
	We set $\Sigmastar={\rm diag}(\Sigma_{11}^*,\ldots,\Sigma_{qq}^*)$ with $q=10$, where $\Sigma_{jj}^*$ was a 10-dimensional square matrix with the above uniform pattern.
\end{enumerate}

\begin{figure}
	\centering
	\includegraphics[height=3.3cm]{figures/ijcai_coef_cor_method.pdf}
	\includegraphics[height=3.3cm]{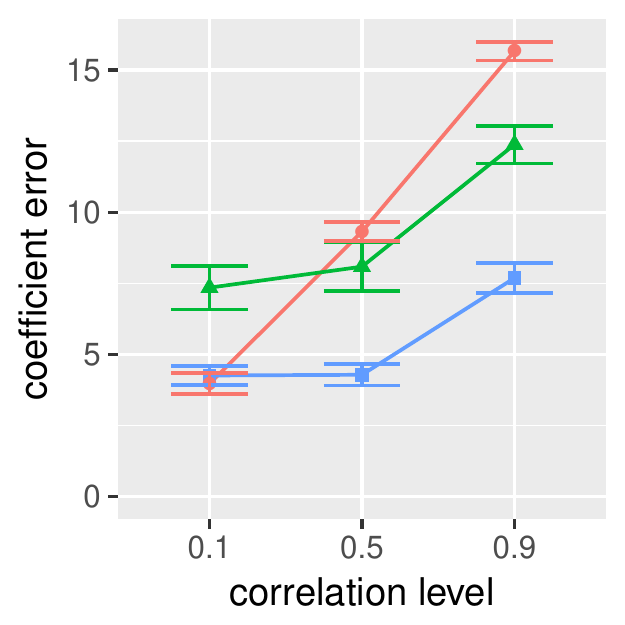}
	\includegraphics[height=3.3cm]{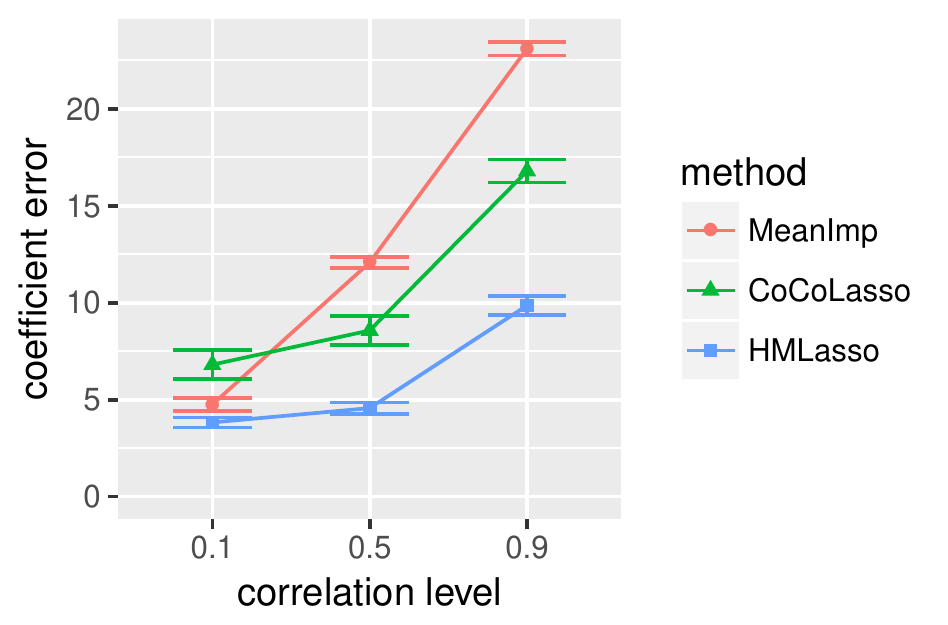}
	\caption{Simulation results with various covariance patterns: uniform (left), autoregression (center), and block (right).}
	\label{fig:cov_pt}
\end{figure}

Figure~\ref{fig:cov_pt} shows the results.
HMLasso outperformed the other methods for almost all covariance patterns and covariance levels.
The mean imputation method was comparable to HMLasso under low covariance conditions, because the shrinkage estimator such as the mean imputation tends to show a good performance when the true covariance is close to zero.
However, the mean imputation deteriorated its estimation under a moderate or high covariance condition.

\subsection{Noise Levels and True Parameters}

We examined three noise levels and three kinds of true parameters.
We set noise levels to ${\rm Var}[\varepsilon] = 0.1^2, 1, 10^2$.
The true parameters $\beta$ were defined as the following.
\begin{enumerate}[(1)]
	\item $\beta_1 = 10, \beta_{11} = -9, \beta_{21} = 8, \beta_{31} = -7, \dots, \beta_{91} = -1$, and $\beta_j = 0$ otherwise.
	\item $\beta_1 = 10, \beta_{2} = -9, \beta_{3} = 8, \beta_{4} = -7, \dots, \beta_{10} = -1$, and $\beta_j = 0$ otherwise.
	\item $\beta_1 = 5, \beta_{11} = -5, \beta_{21} = 5, \beta_{31} = -5, \dots, \beta_{91} = -5$, and $\beta_j = 0$ otherwise.
\end{enumerate}

\begin{figure}
	\centering
	\includegraphics[height=3.3cm]{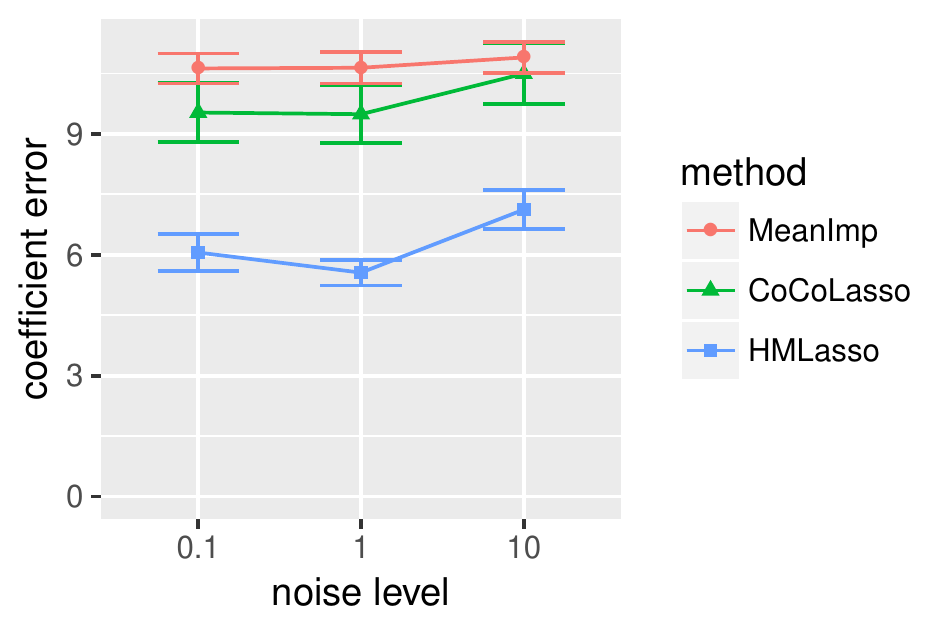}
	\includegraphics[height=3.3cm]{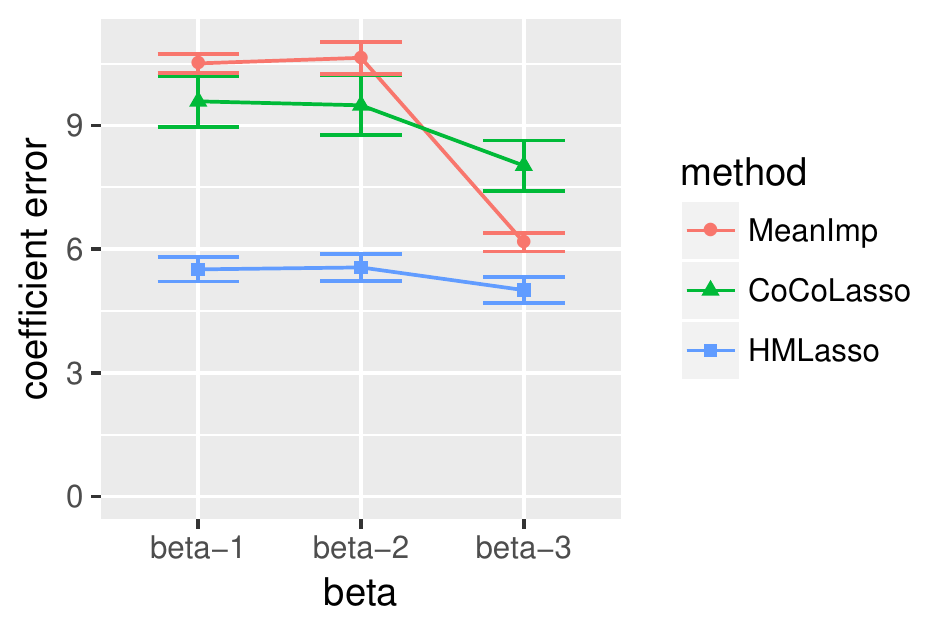}
	\caption{Simulation results with various noise levels (left) and true parameters (right).}
	\label{fig:eps_beta}
\end{figure}

Figure~\ref{fig:eps_beta} shows the results.
HMLasso outperformed the other methods for all noise levels and true parameters.

\newpage

\end{document}